\documentclass[letterpaper, conference]{ieeeconf}
\IEEEoverridecommandlockouts

\usepackage{cite}

\def\BibTeX{{\rm B\kern-.05em{\sc i\kern-.025em b}\kern-.08em
    T\kern-.1667em\lower.7ex\hbox{E}\kern-.125emX}}
    
  \usepackage{amsmath,amssymb,amsfonts,amsthm} 
\usepackage{listings}
\usepackage{mathtools}
\usepackage{commath}
\usepackage{svg}
\usepackage{algorithmic}
\usepackage[ruled,vlined]{algorithm2e}
\usepackage{graphics}
\usepackage{textcomp}
\usepackage[caption=false]{subfig}
\usepackage{framed}
\definecolor{shadecolor}{rgb}{.9,.9,.9}
\usepackage{tcolorbox}
\usepackage{siunitx}
\usepackage{textgreek}
\usepackage{url}
\usepackage{booktabs, tabularx}
\usepackage{multirow}
\usepackage{stackengine}

\usepackage{tikz}
\usetikzlibrary{quotes,angles}
\usetikzlibrary{calc}
\usetikzlibrary{decorations.pathmorphing}

\usepackage{soul}
    \usepackage{todonotes}
\usepackage{tablefootnote}
\usepackage[capitalize]{cleveref}

\newcommand{\crefrangeconjunction}{--}
\crefrangeformat{problem}{Problems #1\crefrangeconjunction #2}
\crefmultiformat{problem}{Problems #1}{ and #1}{, #1}{ and #1}
\crefrangeformat{equation}{#3(#1)#4\crefrangeconjunction#5(#2)#6}

\title{\LARGE\bf Holonomic Control of Arbitrary Configurations of Docked Modboats
}

\author{Zhijie Qiao, Gedaliah Knizhnik, and Mark Yim
\thanks{The authors are with the GRASP Laboratory, University of Pensylvannia, Philadelphia, PA 19104. 
        {\tt\footnotesize zhijie@alumni.upenn.edu}}%
}

\DeclareMathOperator{\sign}{sgn}

\newtheorem{problem}{Problem}

\crefrangeformat{equation}{#3(#1)#4--#5(#2)#6}

\begin{document}
\bstctlcite{MyBSTcontrol} 

\maketitle

\begin{abstract}
The Modboat is a low-cost, underactuated, modular robot capable of surface swimming, docking to other modules, and undocking from them using only a single motor and two passive flippers. Undocking is achieved by causing intentional self-collision between the tails of neighboring modules in certain configurations; this becomes a challenge, however, when collective swimming as one connected component is desirable. Prior work has developed controllers that turn arbitrary configurations of docked Modboats into \textit{steerable} vehicles, but they cannot counteract lateral forces and disturbances. In this work we present a centralized control strategy to create \textit{holonomic} vehicles out of arbitrary configurations of docked Modboats using an iterative potential-field based search. We experimentally demonstrate that our controller performs well and can control surge and sway velocities and yaw angle simultaneously.
\end{abstract}


\section{Introduction} \label{sec:intro}

Aquatic modular self-reconfigurable robotic systems (MSRRs) are of great interest to researchers and industry; they can be used for monitoring ocean environments, performing exploration tasks, and collecting flow information\cite{PALMER2021105805}, while adapting to changing conditions and scales of interest. Conventional wisdom has been that such MSRRs must be built from modules capable of holonomic motion~\cite{Paulos2015, OHara2014, Wang2018DesignVehicle, Wang2020RoboatEnvironments}, which has limited development due to increased complexity. Recent work by the authors, however, has shown that effective aquatic MSRRs can be built from \textit{underactuated} surface-swimming modules~\cite{modboatsOnline,Knizhnik2021,Knizhnik2022,KnizhnikTRO,Knizhnik2021a}. 

\begin{figure}[t]
\centering
\includegraphics[width=\linewidth]{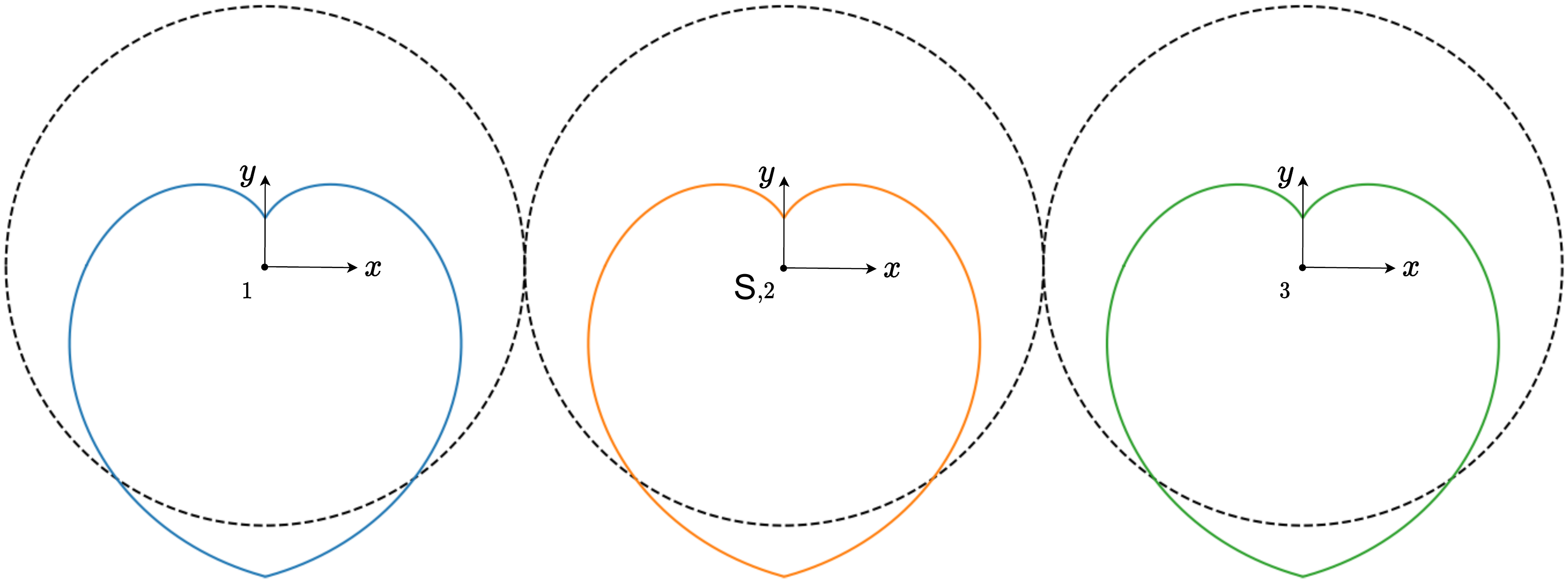}
\caption{Example structure of three docked Modoats in a parallel configuration, with the structure frame given by $S$ and the surge ($y$) and sway ($x$) axes marked. For each boat, the top body is shown in black, and the tail is shown in color, each pointing at $-\pi/2~\si{rad}$. The passive flippers are not shown but connect to the tails and do not protrude from the top body footprint. More details of the Modboat structure can be found in~\cite{modboatsOnline}.}
\label{fig:Struct}
\end{figure}

The underactuated modules used in this prior work --- the Modboats --- use passive flippers and an inertial rotor powered by a single motor to generate thrust and steering~\cite{modboatsOnline}. They are capable of docking and reconfiguration through permanent-magnet based docks and can undock from one another using mechanical self-collision of protruding tails (see~\cref{fig:Struct}), all while using only one motor~\cite{Knizhnik2021}. While this passive docking setup and mechanical undocking method greatly reduces actuation complexity for the system, it also introduces a major constraint when attempting to swim collectively: any collective behavior must constantly avoid self-collisions between neighboring tails, which would cause the docked configuration to disintegrate. 

Prior work addressed this concern by introducing restrictions on the phase and thrust direction allowed for the Modboat modules~\cite{Knizhnik2022,KnizhnikTRO}, which are technically capable of thrusting in any direction~\cite{Knizhnik2021a}. By limiting the thrust direction to the surge axis of the configuration and requiring in-phase swimming, these approaches were able to guarantee no unintentional self-collisions and maintain \textit{steerability} for any arbitrary structure~\cite{Knizhnik2022,KnizhnikTRO}. But the resulting controllers could not produce force along the configurations' sway axes, which made them highly susceptible to noise and external disturbances. It is also reasonable to suspect that a structure of $N\geq3$ modules should be holonomic in the plane, given a reasonable control law, but thus far such a controller has not been developed due to the complexity of the collision constraint~\cite{KnizhnikTRO}.  

In this work, we propose a \textbf{holonomic} collective control approach for an arbitrary configuration of docked Modboats, by using an iterative potential field search to find collision-free desirable movements. While iterative path planning has been widely adopted in autonomous robot navigation~\cite{9636071,YAO2019455,6094881,5717393}, little work has been done to explore its use in the avoidance of internal collision constraints. Our controller uses such an approach to relax the assumption in~\cite{Knizhnik2022,KnizhnikTRO} of thrust only along the surge axis, while using a collision checker to avoid any unintended undocking. 

The rest of this work is organized as follows: in~\cref{sec:Methodology} we formulate the problem and present an approach for finding collision-free motions that generate desired forces. \cref{sec:Control} presents the control approach for determining desired forces for given motions, and~\cref{sec:experiments} presents the experimental evaluation of our approach. The results are discussed in~\cref{sec:Discussion}.

\section{Methodology} \label{sec:Methodology}

Consider a configuration of docked Modboat modules labeled $i \in [1,N]$. To move, the modules collectively perform a series of \textbf{swim cycles}, in which each module executes the waveform given in~\eqref{eq:modboatWaveform}, where $\phi_i(t)$ is the motor angle of module $i$ and $\phi_{0,i}$, $A_i$, and $T$ are the centerline, amplitude, and period of the oscillation, respectively. The period is constant for all modules for concurrency\footnote{In this work we use $T=1.5~\si{s}$, which has been empirically determined to be an effective period for the system~\cite{Knizhnik2020a,KnizhnikTRO}, although the methodology is applicable to any period.}, while the centerline and amplitude vary for each module for control. Parameters are determined at the beginning of each cycle and executed for a single period. Since the final pose of one cycle and the initial pose of the next cycle may not be coincident, an additional \textbf{transition cycle} is allowed between swim cycles to prepare, after which point the process repeats.
\begin{equation} \label{eq:modboatWaveform}
    \phi_i(t) = \phi_{0,i} + A_i\cos{\left ( 2\pi t/T \right )}, \quad t \in [0, T)
\end{equation}
\begin{equation} \label{eq:forceFromAmplitude}
    F(A) = 0.022 \abs{A} - 0.019, \quad \abs{A} \in [0.9,2.6]
\end{equation}

In prior work we have shown that the use of waveform~\eqref{eq:modboatWaveform} results in a linear relationship between force and amplitude, given in~\eqref{eq:forceFromAmplitude} for a period of $1.5~\si{s}$\cite{Knizhnik2022,KnizhnikTRO}. Unlike in prior work~\cite{Knizhnik2022,KnizhnikTRO}, however, in this work we allow $\phi_0$ to take on any value. We also allow amplitude to take on both positive and negative values for numerical flexibility; negative amplitudes result in a phase shift of $\pi~\si{rad}$, but do not affect the generated force, as per~\eqref{eq:forceFromAmplitude}.

\begin{figure}[t]
  \centering
  \includegraphics[width=0.7\linewidth]{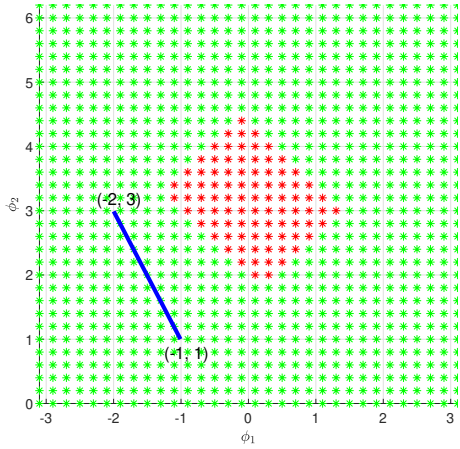}
  \caption{Numerically simulated phase-space with collision region, with $\phi_1$ being the left boat and $\phi_2$ being the right boat. Collided configurations are shown in red, and free space in green, and top-body neighbors display an offset version of this relationship. The blue line represents a sample motion, with boat $1$ having $(\phi_0,A) = (-1.5,0.5)$, and boat $2$ having $(\phi_0,A) = (2.0, 1.0)$}
  \label{fig:CollisionRegion}
\end{figure}

\begin{problem}[Valid Swim Cycles]\label{prob:mainProb}
Given a configuration of docked Modboats $i\in[1,N]$ and a set of desired forces $$\vec{F}_{des} = \begin{bmatrix} F_{x,des} & F_{y,des} & \tau_{des} \end{bmatrix}^T$$ find a set $\Phi$ of pairs $(\phi_0,A)_i$ for $i\in[1,N]$ such that the generated forces $$\vec{F} = \begin{bmatrix} F_{x} && F_{y} && \tau \end{bmatrix}^T $$ equal (as closely as possible) the desired forces, i.e. $\vec{F} = \vec{F}_{des}$, while avoiding tail collisions between neighboring modules.
\end{problem}

\begin{problem}[Valid Transition Cycles]\label{prob:secondProb}
Given multiple sets of motion pairs $\Phi_k$, $k \in \mathbb{Z}^+$, where $k$ represents the swim cycle for which the solution is used, find a set of motions $\Psi$ to transition from the final pose in $\Phi_k$ to the initial pose in $\Phi_{k+1}$ while avoid tail collisions between neighboring modules.
\end{problem}

The goal of this work, then is to find solutions to~\cref{prob:mainProb,prob:secondProb} under these conditions, i.e. to find a valid set of swim and transition cycles that generate a desired set of forces without causing internal collisions. This will allow the configuration of docked Modboats to function as a \textbf{holonomic} vehicle. 


\subsection{Collision Region} \label{sec:collisionRegion}

The first step in solving~\cref{prob:mainProb} is to identify what set of poses causes a collision between neighbors. We can construct a graph of this collision region by numerically solving for the intersection of the tails for any pair of boats, as described in~\cite{KnizhnikTRO}. The resulting graph is shown in~\cref{fig:CollisionRegion} for a single pair of boats; in this phase space any motion of the two neighboring boats is a straight line, as proven in~\cite{KnizhnikTRO}.

\begin{figure}[t]
\centering
\includegraphics[width=0.75\linewidth]{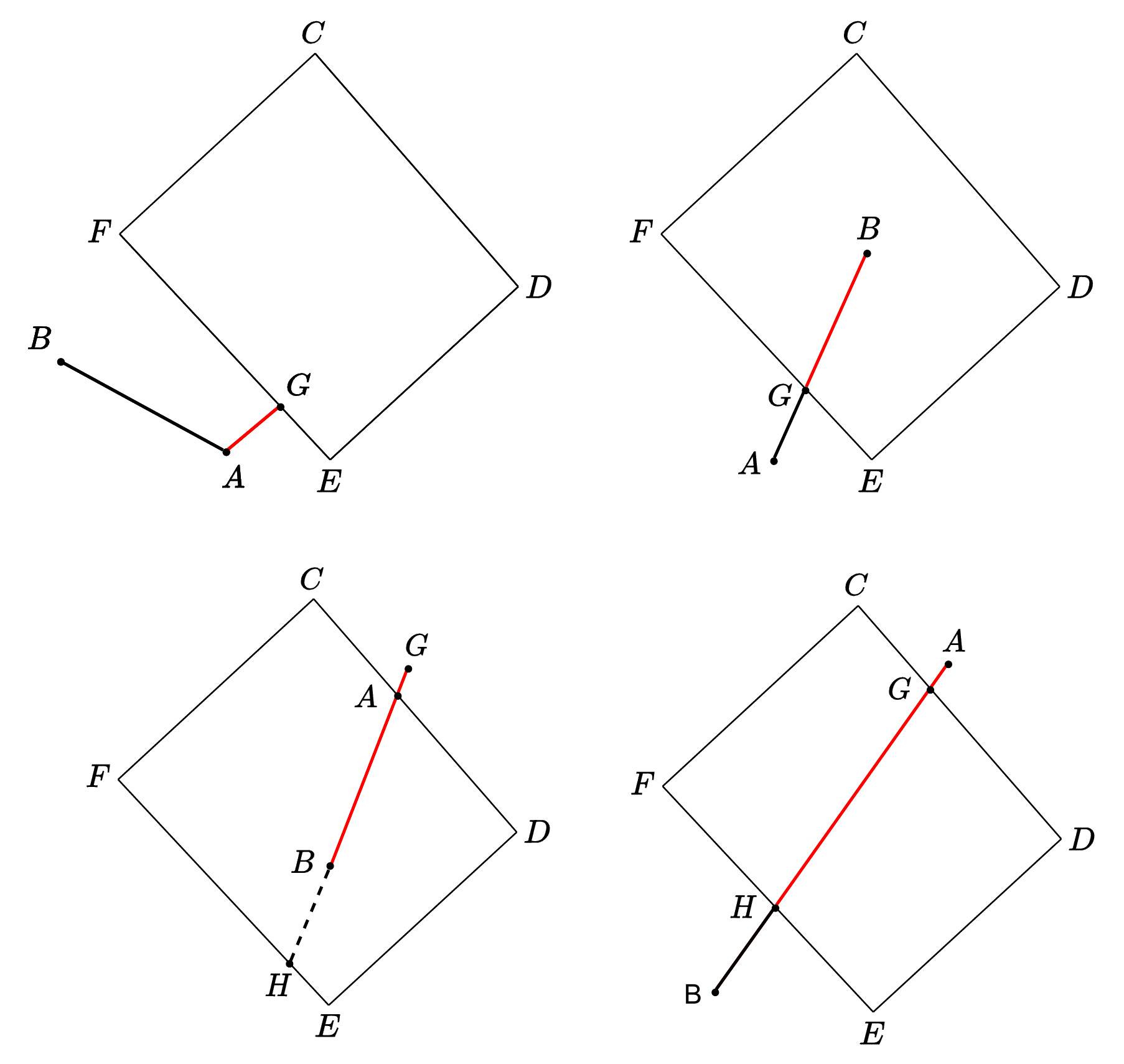}
\caption{Distance to Collision (DoC) analysis for four different intersecting cases, using the collision region shown in~\cref{fig:CollisionRegion}. $AB$ is the phase-space trajectory and $CDEF$ is the collision region. The DoC is shown in red in each case.}
\label{fig:DoC}
\end{figure}

The critical question is how to define the distance to collision (DoC) given the phase-space representation in~\cref{fig:CollisionRegion} and linear trajectories. For any pair of boats, we can define the DoC as $DoC = D(\phi_i,A_i, \phi_j,A_j)$. Alternatively, given a line $AB$ representing a trajectory in phase-space and a polygon with boundary $CDEF$ representing the collision region, we propose to define DoC as follows and as illustrated in~\cref{fig:DoC}:

\begin{enumerate}
   \item If $AB$ is completely outside $CDEF$, the DoC is the minimum distance from $AB$ to $CDEF$.
   \item If $AB$ intersects $CDEF$ on point $G$ with $B$ interior to $CDEF$, the DoC is $-|BG|$.
   \item If $AB$ is inside $CDEF$ with extensions intersecting $CDEF$ on $G$ and $H$, the DoC is $- \left ( |AB| + \min[|AG|, |BH|] \right )$.
   \item If $AB$ intersects $CDEF$ on $G$ and $H$, the DoC is $- \left (|GH| + \min[|AG|, |BH|] \right )$.
\end{enumerate}

In summary, the DoC captures the distance along the phase-space trajectory to move into or out of the collision region. While more optimal strategies (i.e. moving sideways) exist, this strategy is physically meaningful in the context of the Modboats. Computing the DoC is also expensive, since the calculation needs to be performed for each pair of boats at every step in the solution process, so we precompute a DoC table with a discretization of $0.1~\si{rad}$ for the centerline and amplitude. 

Note that in practice~\cref{fig:CollisionRegion} must be extended to $[-2\pi, 2\pi]$ to account for angle wrapping, which creates several identical but shifted collision regions. In this case the DoC must be calculated for every collision region and the most conservative value is used.


\subsection{Attractive Field} \label{sec:attractiveField}

Given the collision-space representation developed in~\cref{sec:collisionRegion}, we begin to solve~\cref{prob:mainProb} by creating an attractive potential to drive the values for generated structural forces $\vec{F}$ to their desired values $\vec{F}_{des}$, which can be computed based on the gradient of an error term. Let our error vector be given by $\vec{e}$ in~\eqref{eq:errorVec}, and our weight vector be given by $\vec{w}$ in~\eqref{eq:weightVec}, which accounts for the relative unit scale of force and torque (we have heuristically determined that $[\begin{matrix} 1 & 1 & 10 \end{matrix}]^T$ is effective). Then the error in generated forces is given by $E = \vec{e} \cdot \vec{w}$.
\begin{equation}\label{eq:errorVec}
    \vec{e} = \vec{F}_{des} - \vec{F}
\end{equation}
\begin{equation}\label{eq:weightVec}
    \vec{w} = \begin{bmatrix}
        w_x & w_y & w_\tau
    \end{bmatrix}^T
\end{equation}

Eq.~\eqref{eq:modboatWaveform} is parameterized by $\phi_{0,i}$ and $A_i$ for each boat, or $\vec{\phi}_0$ and $\vec{A}$ for all boats. Then~\eqref{eq:gradPhi} and~\eqref{eq:gradA} give the gradient in terms of those quantities.
\begin{align}
    \nabla_{\vec{\phi}_0} E &=
        -\left (\nabla_{\vec{\phi}_0}  \vec{F} \right )  \vec{w} \label{eq:gradPhi} \\
    \nabla_{\vec{A}\hphantom{_|}} E &=
        -\left ( \nabla_{\vec{A}\hphantom{_|}} \vec{F} \right )  \vec{w} \label{eq:gradA}
\end{align}

From~\eqref{eq:forceFromAmplitude} and~\cref{fig:Struct}, we can define the forces produced by each boat as in~\eqref{eq:forceDecomposition}, which combine to form the configuration's generated forces as $\vec{F} = \sum_{i} \vec{F}_i$. Note that the generated force for each boat points opposite the centerline direction of the tail tip, and $F_i = F(A_i)$ from~\eqref{eq:forceFromAmplitude}.
\begin{equation} \label{eq:forceDecomposition}
    \vec{F}_{i} = \begin{bmatrix}
        F_{x}\\
        F_{y}\\
        \tau
    \end{bmatrix}_i = 
    \begin{bmatrix}
        -F_i \cos(\phi_{0,i})\\
        -F_i \sin(\phi_{0,i}) \\
        -F_i \sin(\phi_{0,i}) x_i + F_i \cos(\phi_{0,i}) y_i
    \end{bmatrix}
\end{equation}

 \begin{figure}[t]
  \centering
  \fontsize{30}{12}\selectfont
  \includegraphics[width=\linewidth]{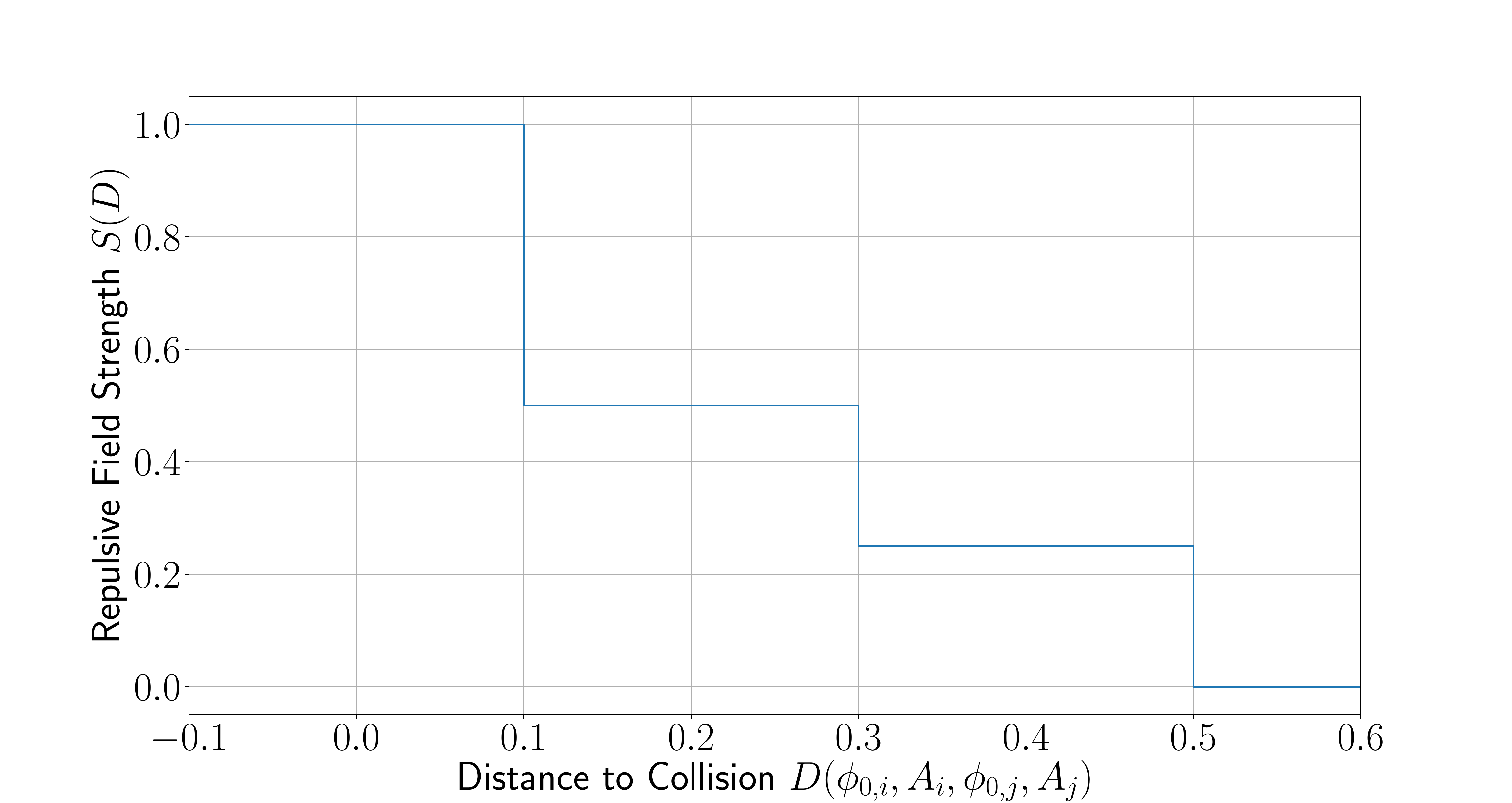}
  \caption{Repulsive field strength segmented function vs.distance to collision. This stepped function determines the strength of the repulsive field and prioritizes the worst collisions.}
  \label{fig:universe}
\end{figure}

The gradient of the force produced by each boat can be taken as in~\eqref{eq:gradPhi} with respect to its centerline, and~\eqref{eq:gradA} with respect to its amplitude. These gradients then form the rows of the gradients in~\eqref{eq:gradPhi} and~\eqref{eq:gradA} with respect to $\vec{\phi}_0$ and $\vec{A}$, respectively. 
\begin{align}
    \nabla_{\phi_{0,i}} \vec{F}_{i} &= \begin{bmatrix}
        \hphantom{-}F_i \sin(\phi_{0,i})\\
        -F_i \cos(\phi_{0,i})\\
        -F_i \cos(\phi_{0,i})x_i - F_i \sin(\phi_{0,i}) y_i \end{bmatrix}^T \label{eq:gradFPhi}\\
    \nabla_{A_i} \vec{F}_{i} &=
        \begin{bmatrix}
        -0.022 \cos(\phi_{0,i})\\
        -0.022 \sin(\phi_{0,i})\\
        -0.022 \left [\sin(\phi_{0,i})x_i -\cos(\phi_{0,i}) y_i \right ]
    \end{bmatrix}^T \label{eq:gradFA}
\end{align}

Eqs.~\eqref{eq:gradPhi} and~\eqref{eq:gradA} can then be used --- with values obtained from~\eqref{eq:gradFPhi} and~\eqref{eq:gradFA} --- to obtain the desired step direction from the attractive field as in~\eqref{eq:step}. However, since the size of this step may be arbitrarily small, we take a fixed size step in the \textit{direction} given by the gradient, where we use the $\sign$ function rather than taking the unit vector to account for the \textit{discretized} phase space.
\begin{align} \label{eq:step}
\begin{split}
    \vec{d}_{\vec{\phi}_0} &= \Delta d \hphantom{|}\sign{(\nabla_{\vec{\phi}_{0}} E )} \\
    \vec{d}_{\vec{A}\hphantom{_|}} &= \Delta d \hphantom{|}\sign{(\nabla_{\vec{A}\hphantom{_|}} E)}
\end{split}
\end{align}


\subsection{Repulsive Field} \label{sec:repulsiveField}

To solve~\cref{prob:mainProb} we also need a repulsive field to drive each Modboat's tail away from collisions with any of its neighbors, which occur at motor angles given by~\cref{fig:CollisionRegion}. Because of the non-standard nature of distance to collision in this scenario, we take the following approach: for each boat, let $D_0$ represent the DoC at the current state, while $D_\phi$ and $D_A$ represent the DoC after updating $\phi_{0,i}$ and $A_i$ while neighboring parameters (subscript $j$) remain constant, as in~\eqref{eq:docUpdate}.

\begin{equation}\label{eq:docUpdate}
    \begin{bmatrix}
        D_0\\
        D_\phi\\
        D_A
    \end{bmatrix}_{i,j} = 
\begin{bmatrix}
    D(\phi_{0,i} \hphantom{xx   d_{\phi_{0,i}}} , A_i \hphantom{xx d_{A,i}}          , \phi_{0,j}, A_j)\\
    D(\phi_{0,i} + d_{\phi_{0,i}}, A_i \hphantom{xx  d_{A,i}} , \phi_{0,j}, A_j)\\
    D(\phi_{0,i} \hphantom{xx   d_{\phi_{0,i}}}     , A_i + d_{A,i}, \phi_{0,j}, A_j) 
    \end{bmatrix}
\end{equation}

The repulsive field strength $U_i$ is constructed in~\eqref{eq:repulsive} for each boat using a segmented expression based on the DoC, as shown in~\cref{fig:universe}, which has been shown to work well in our experiments, and its sign is adjusted based on whether the attractive field step results in ``more'' or ``less'' collision. 
\begin{equation}\label{eq:repulsive}
    \begin{bmatrix}
        U_\phi \\
        U_A
    \end{bmatrix}_{i} = \sum_{j} \left (
    \begin{bmatrix}
        \sign{( D_\phi - D_0)} \\
        \sign{(D_A - D_0)}
    \end{bmatrix} S(D_0) \right )_{i,j}
\end{equation}

The repulsive field outputs in~\eqref{eq:repulsive} are calculated by summing over all occupied neighbor sites $j$ (maximum 4). If $U_i$ is positive, the update in~\eqref{eq:step} is performed for boat $i$; otherwise, no update is performed. This allows each boat to prioritize the worst collision case and reach a balance among all its neighbors.

\subsection{Overall Procedure}\label{sec:methodology:overallProcedure}

\begin{algorithm}[t]
\caption{Potential field algorithm for generating collision free solutions to~\cref{prob:mainProb}.}\label{alg:potentialField}
\For{$n_{epoch} \in [1,N]$}{
    \For{$i \in [1,N_{1}]$}{
        Apply \textbf{attractive} field\;
    }
    \For{$i \in [N_{1},N_{2}]$}{
        Apply \textbf{ attractive} field if \textbf{repulsive} field $>$ 0\;
    }
    \For{$i \in [N_{2},N_{3}]$}{
        Apply -\textbf{attractive} field if \textbf{repulsive} field $<$ 0\;
    }
}
\end{algorithm}
The attractive field in~\cref{sec:attractiveField} and repulsive field in~\cref{sec:repulsiveField} together should be sufficient to solve~\cref{prob:mainProb}. The repulsive field formulation in~\cref{sec:repulsiveField} has two issues, however: (a) it does not \textit{guarantee} the final solution will be collision free, and (b) the Modboats may be unable to traverse through collided phase-space to reach a desired solution. To address these issues, we apply a three-stage iterative approach to solving~\cref{prob:mainProb} as presented in~\cref{alg:potentialField}:
\begin{enumerate}
    \item Applying the attractive field alone prioritizes finding \textit{a} solution to the desired forces.
    \item Applying the attractive field and repulsive field together searches for a \textit{valid} solution.
    \item Applying only the repulsive field alone prioritizes ensuring the solution is valid.
\end{enumerate}

This approach is quite effective, as shown in~\cref{sec:experiments}, although it still does not guarantee the final solution is fully collision free or globally optimal. Since collision avoidance is critical, the final solution \textit{used} is the most optimal collision-free solution found along the way. 

When applied to all modules $i \in [1,N]$ concurrently, \cref{alg:potentialField} produces a collision-free set $\Phi$ of $(\phi_0,A)_i$ $\forall i \in [1,N]$, solving~\cref{prob:mainProb}. Each Modboat then executes~\eqref{eq:modboatWaveform} with those parameters for a single \textbf{swim cycle} (i.e. a single period), before repeating the process for the next cycle and set of desired forces.



\subsection{Transition Solver} \label{sec:Transition Solver}

 \begin{figure}[t]
  \centering
  \includegraphics[width=0.75\linewidth]{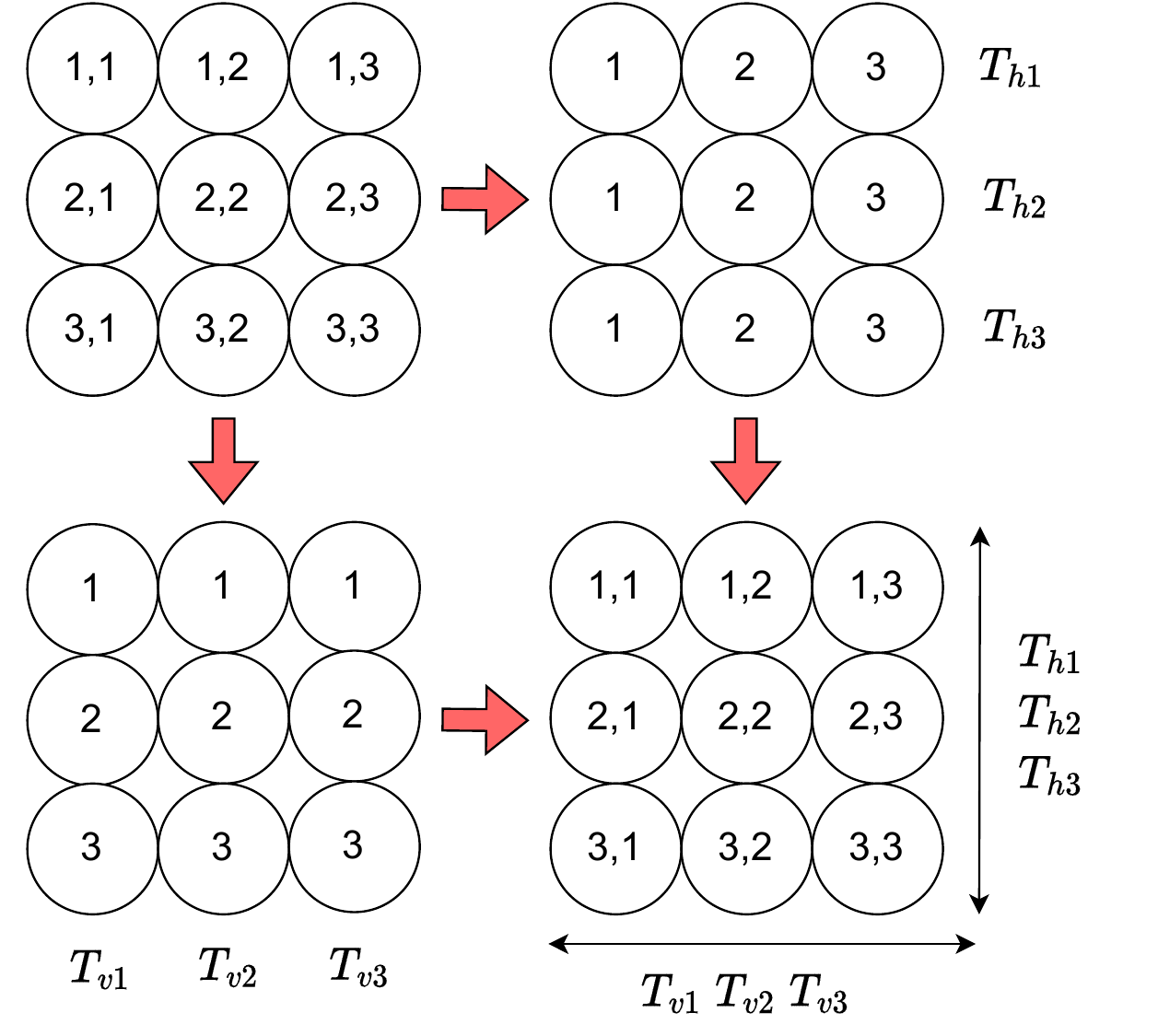}
  \caption{To make finding valid transition sets more tractable, we decompose the structure into horizontal and vertical sub-problems. For each sub-problem, only interactions in the given direction are considered, and solution \textit{sets} are found via AC~\cite{MACKWORTH197799} for each boat. The full collision-free solution is the intersection of the solution sets for each boat in the sub-problems.}
  \label{fig:transitionAC}
\end{figure}

Although the procedure in~\cref{sec:methodology:overallProcedure} solves~\cref{prob:mainProb} and produces collision free movements, when used to generate a sequence of movements $\Phi_k$ for $k\in\mathbb{Z}^+$ it does not guarantee that the transition from the \textit{last} position of one movement to the \textit{start} of the next is itself collision free. An additional transition solver is needed to solve~\cref{prob:secondProb} and find a collision free set of transition paths $\Psi$. We assume, for ease of computation, that transitions take the entirety of $t_{trans}$ for all boats, and occur at a constant speed.

Any Modboat can take either a clockwise or counterclockwise path from the last position of its previous cycle to the first position of its next. For any pair of modules, then, there are four possible transitions, each of which is a line in the phase-space in~\cref{fig:CollisionRegion} and can be simply checked for collision. Of these, at least one is likely to be collision-free, but to increase the number of available options we also consider negating the amplitude of all boats in the next cycle, which shifts the start locations of the next cycle but does not otherwise affect the solution. This set of possible transitions is then run through the arc consistency (AC) algorithm~\cite{MACKWORTH197799} to find a valid collision-free transition for each boat pair.

Because Modboat configurations are two-dimensional there many interactions between neighbors, since diagonal neighbors --- while not directly neighbors --- influence each other through shared connections. To make the search space more tractable, we split the spatial constraint by considering horizontal interactions and vertical interactions as two separate problems, as in~\cref{fig:transitionAC}. When AC is run on each row (column) of the horizontal (vertical) sub-problem, it generates a \textit{set} of valid transitions for each boat. The full solution is then generated by intersecting the solution sets for each boat from the horizontal and vertical sub-solutions, which recovers the full spatial constraint. 

In our testing with simulated random inputs and prior locations, a valid transition set exists for all boats in over $99\%$ of cases for square configurations of Modboats with 2--5 boats to a side. However, since the number of possible sets to evaluate scales exponentially with the number of boats, this approach becomes intractable for larger structures; a more efficient strategy will need to be developed for larger configurations, but is left to future work.


A critical thing to note is that the transition requires a finite $t_{trans}$ between cycles of~\eqref{eq:modboatWaveform}, which means that decisions about generated forces are made at intervals of $T+t_{trans}$. Using a small $t_{trans}$ maintains the responsiveness of the overall controller, but the fast transitions introduce unwanted dynamic disturbances. Using a large $t_{trans}$ minimizes the dynamic disturbances, but slows the response of the overall controller. We minimize this impact by selecting the solution set that results in the minimum overall distance travelled, but we also expect most transitions to be small during normal operation, since the control generated in~\cref{sec:Control} should be relatively continuous unless sharp maneuvers are needed.


\section{Control} \label{sec:Control}

The methodology of~\cref{sec:Methodology} provides the parameters for~\eqref{eq:modboatWaveform} --- namely $\phi_0$ and $A$ --- for each boat $i$ in a structure given its shape and a set of desired values $\vec{F}_{des}$. To determine these desired values and implement holonomic control for the configuration as a whole, PID control is applied to the equations of motion as derived in~\cite{Knizhnik2022}.

Control for a desired yaw angle $\Theta$ is given in~\cref{eq:yawError,eq:alpha,eq:torque}, where $\Omega$ is the observed angular velocity of the structure, $T$ is the period of~\eqref{eq:modboatWaveform}, $I$ is the Modboat configuration's moment of inertia, and $C_R$ is a drag coefficient~\cite{KnizhnikTRO}. Note that~\eqref{eq:yawError} includes a prediction of the yaw at the end of the current cycle, which accounts for the delay introduced by discrete control during sharp yaw maneuvers.
\begin{equation} \label{eq:yawError}
    e_{\Theta} = \Theta_{des} - \left (\Theta_{obs} + \Omega T \right )
\end{equation}
\begin{equation} \label{eq:alpha}
    \alpha = K_{p,\Theta}e_{\Theta} + K_{d\Theta}\frac{de_{\Theta}}{dt}
\end{equation}
\begin{equation} \label{eq:torque}
\tau_{des} = I\alpha + C_R|\Omega|\Omega
\end{equation}

Velocity control in~\cite{Knizhnik2022,KnizhnikTRO} --- which presented \textit{steerable} vehicles --- considered the desired velocity as a surge (i.e. body-fixed frame) velocity value. Since the method presented in this paper results in a \textit{holonomic} vehicle, we consider the desired values as $\vec{v}_{des} = [\begin{matrix} v_{x,des} & v_{y,des} \end{matrix}]^T$ expressed in the world frame.
Just as in~\cite{Knizhnik2022} an artificial linear acceleration is computed based on the observed error and then integrated into the commanded velocity $\vec{v}_c$, as in~\cref{eq:errVel,eq:acc,eq:velIntegral}.
\begin{equation}\label{eq:errVel}
    \vec{e}_v = \vec{v}_{des} - \vec{v}_{obs}
\end{equation}
\begin{equation} \label{eq:acc}
    \vec{a}_y = K_{p,v}\vec{e}_v + K_{d,v}\frac{d\vec{e}_v}{dt}
\end{equation}
\begin{equation} \label{eq:velIntegral}
    \vec{v}_c = \vec{v}_{des} + (\gamma^{n-1}\sum_{i=0}^{n-1}\vec{a}_y + \vec{a}_y) T
\end{equation} 

A diminishing coefficient $\gamma < 1$ is added to~\eqref{eq:acc} to prevent control lag due to excessive error accumulation. The commanded velocity is then converted to desired force values  using the quadratic drag relationship~\cite{Knizhnik2022} and converted to the boat frame using~\eqref{eq:forceFromCommand}. $\vec{F}_{des}$ can then be constructed from~\eqref{eq:forceFromCommand} and~\eqref{eq:torque} and decomposed into $\vec{\phi}_0$ and $\vec{A}$ using the methodology in~\cref{sec:Methodology}.
\begin{equation} \label{eq:forceFromCommand}
    \begin{bmatrix} F_{x,des} \\ F_{y,des} \end{bmatrix} = C_L\begin{bmatrix}
        \hphantom{-}\sin(\Theta_{obs}) & \cos(\Theta_{obs}) \\
        -\cos(\Theta_{obs}) & \sin(\Theta_{obs})
    \end{bmatrix}\left(\vec{v}_c \odot \vec{v}_c \right )
\end{equation}


\section{Experiments} \label{sec:experiments}

Experiments were conducted in a $4.5~\si{m} \times 3.0~\si{m} \times 1.2~\si{m}$ tank of still water equipped with an OptiTrack motion capture system that provides the real-time position, velocity, and orientation data for each boat at $120~\si{Hz}$. The control methodology in~\cref{sec:Methodology,sec:Control} was computed in Python on an offboard PC and, the resulting parameters for~\eqref{eq:modboatWaveform} and transitions were sent to each Modboat via WiFi at the beginning of each cycle.

\begin{figure}[t]
     \centering
     \subfloat[\label{fig:tests:test1}]{
        \includegraphics[width=0.45\linewidth]{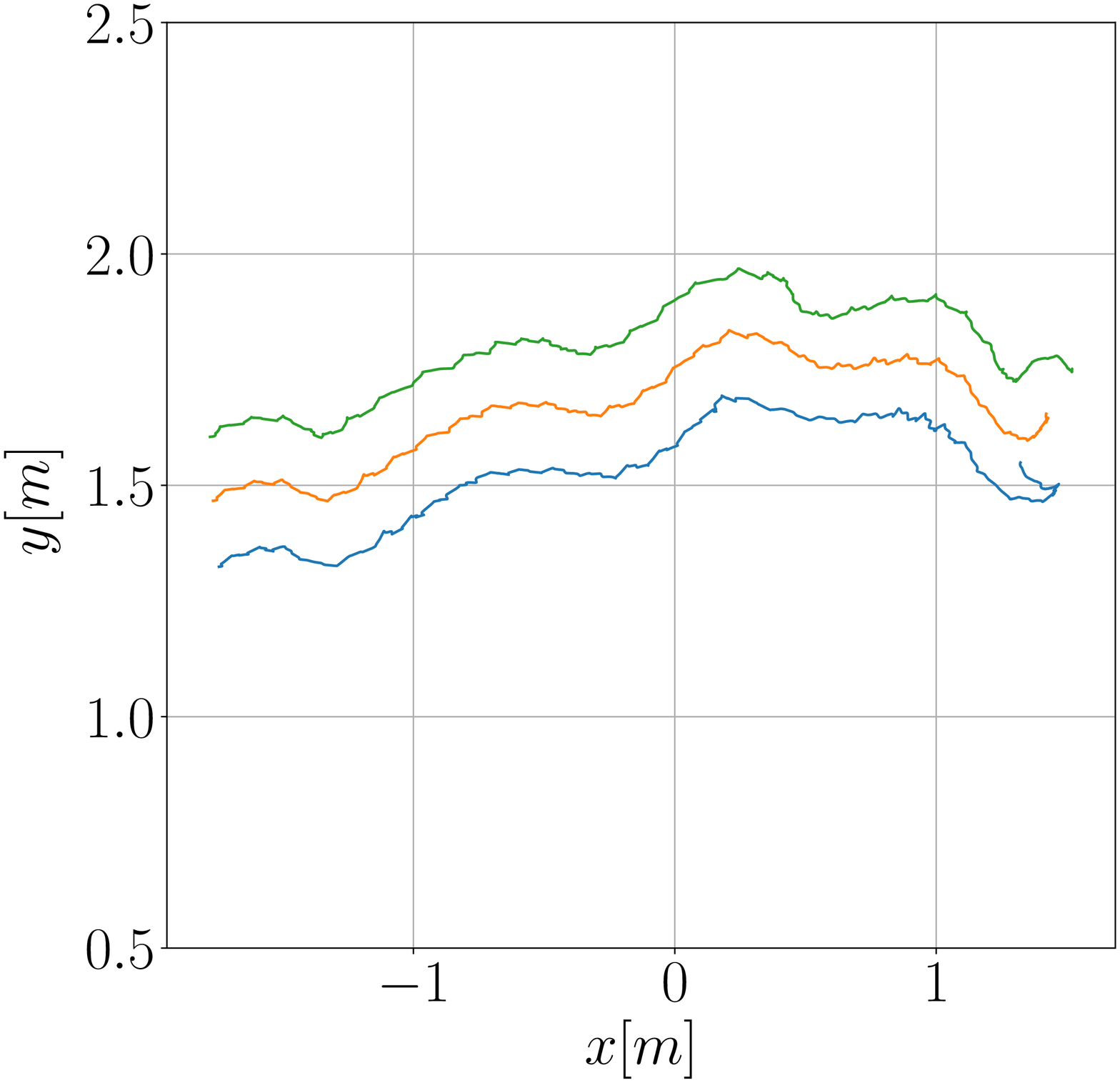}
      } \hfill 
      \subfloat[\label{fig:tests:test2}]{
        \includegraphics[width=0.45\linewidth]{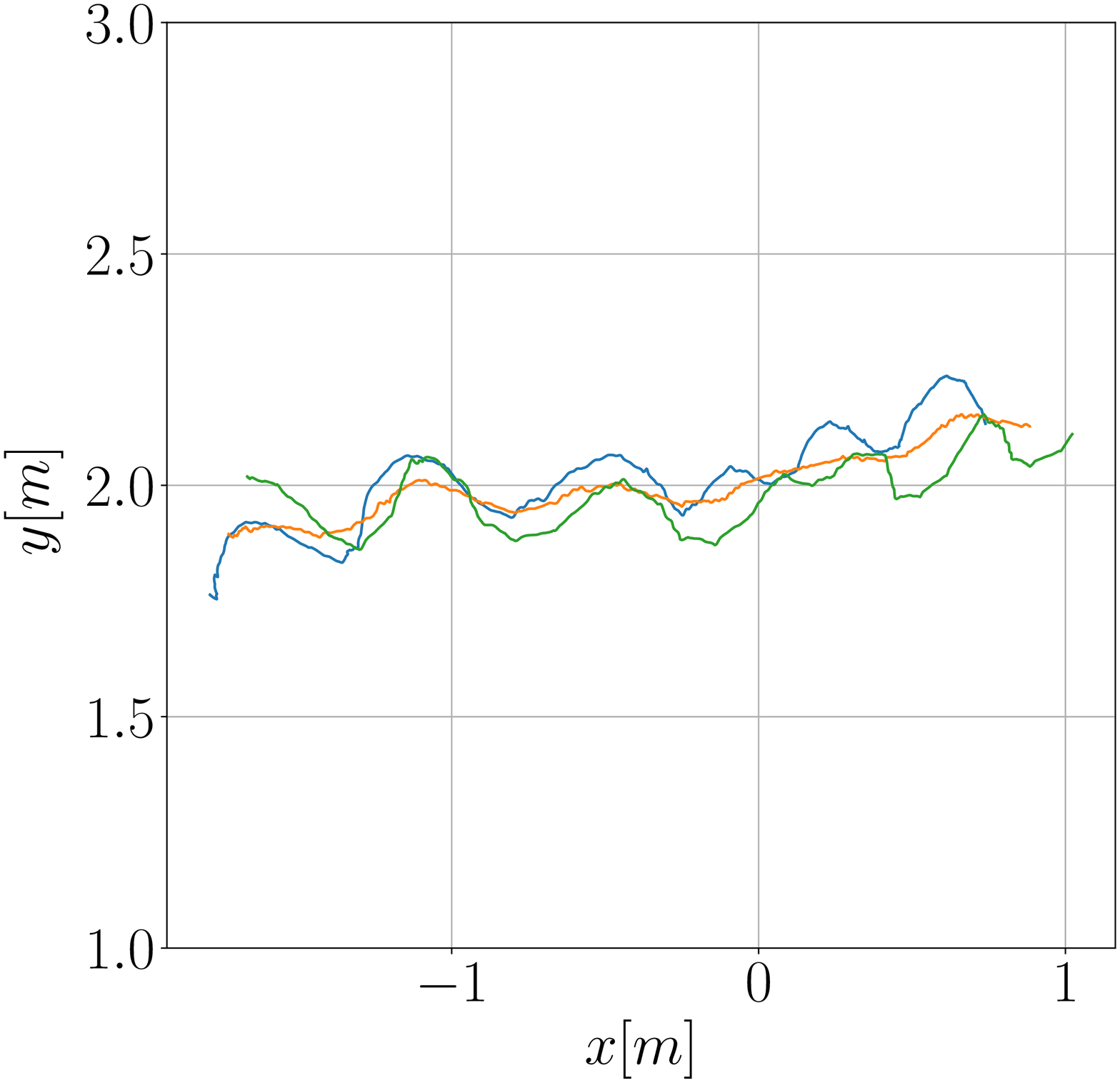}
      } \hfill 
      \subfloat[\label{fig:tests:test3}]{
        \includegraphics[width=0.45\linewidth]{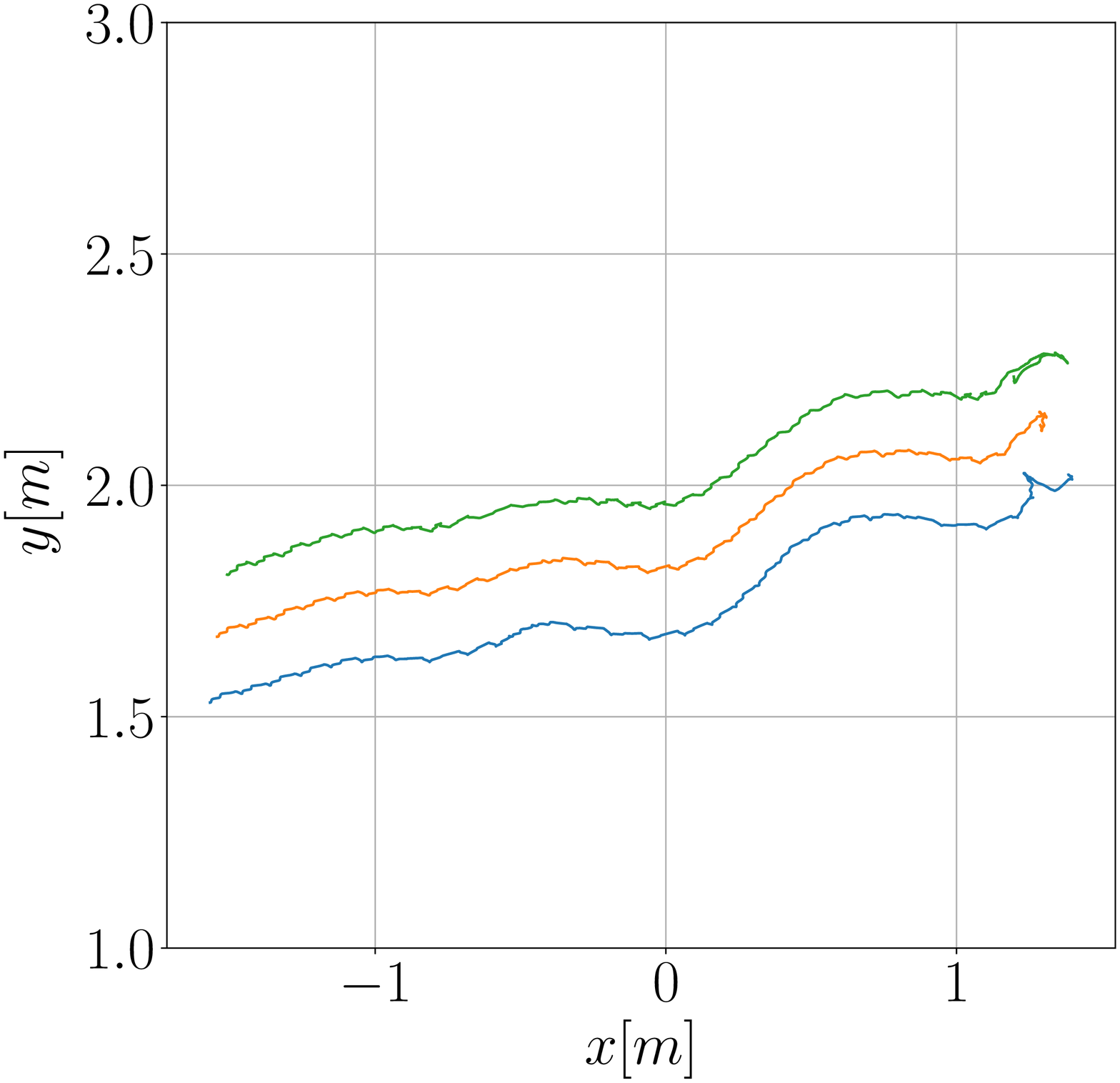}
      } \hfill 
      \subfloat[\label{fig:tests:test4}]{
        \includegraphics[width=0.45\linewidth]{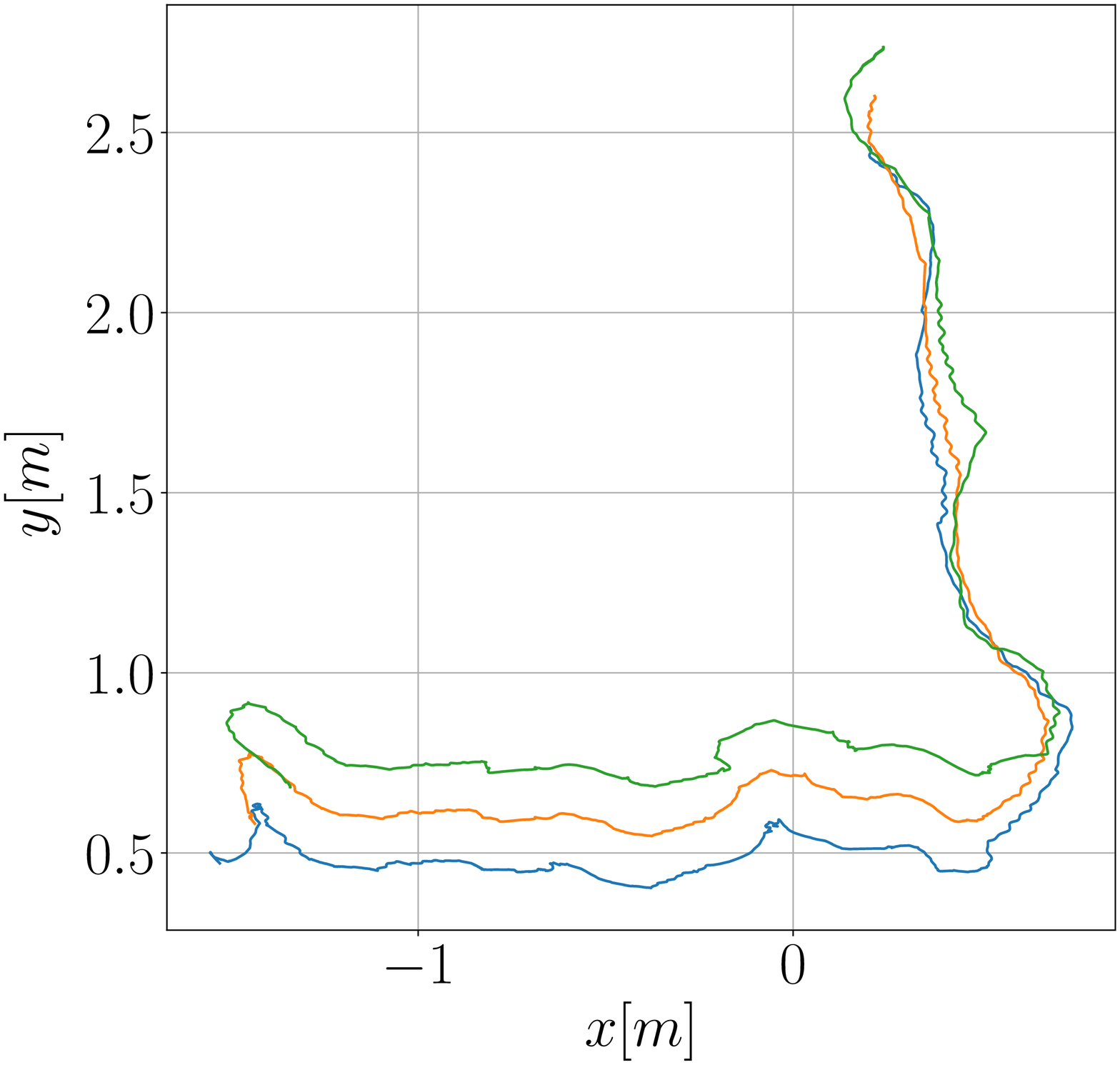}
      } 
      \caption{Experimental trajectories for the four preliminary tests. $[\begin{matrix} v_{x} & v_{y} & \Theta \end{matrix}]_{des} =$ (a) $[\begin{matrix} 0.04~\si{m/s} & 0~\si{m/s} & \pi/2~\si{rad} \end{matrix}]$ (b) $[\begin{matrix} 0.04~\si{m/s} & 0~\si{m/s} & 0~\si{rad} \end{matrix}]$ (c) $[\begin{matrix} 0.03~\si{m/s} & 0.01~\si{m/s} & \pi/2~\si{rad} \end{matrix}]$. (d) $[\begin{matrix} 0.04~\si{m/s} & 0~\si{m/s} & \pi/2~\si{rad} \end{matrix}]$, then $[\begin{matrix} 0~\si{m/s} & 0.04~\si{m/s} & \pi/2~\si{rad} \end{matrix}]$. The COM pose is shown in orange.}
    \label{fig:TESTS}
\end{figure}

Experimental evaluation of the methodology presented in this work is ongoing, but preliminary results show great potential. The following four evaluations have so far been conducted using three boats in a parallel configuration, and the results shown in~\cref{fig:TESTS}:

\begin{enumerate}
    \item $[\begin{matrix} v_{x} & v_{y} & \Theta \end{matrix}]_{des} = [\begin{matrix} 0.04~\si{m/s} & 0~\si{m/s} & \pi/2~\si{rad} \end{matrix}]$ for a duration of $90~\si{s}$.
    \item $[\begin{matrix} v_{x} & v_{y} & \Theta \end{matrix}]_{des} = [\begin{matrix} 0.04~\si{m/s} & 0~\si{m/s} & 0~\si{rad} \end{matrix}]$ for a duration of $90~\si{s}$.
    \item $[\begin{matrix} v_{x} & v_{y} & \Theta \end{matrix}]_{des} = [\begin{matrix} 0.03~\si{m/s} & 0.01~\si{m/s} & \pi/2~\si{rad} \end{matrix}]$ for a duration of $90~\si{s}$.
    \item $[\begin{matrix} v_{x} & v_{y} & \Theta \end{matrix}]_{des} = [\begin{matrix} 0.04~\si{m/s} & 0~\si{m/s} & \pi/2~\si{rad} \end{matrix}]$ for a duration of $60~\si{s}$, then $[\begin{matrix} v_{x} & v_{y} & \Theta \end{matrix}]_{des} = [\begin{matrix} 0~\si{m/s} & 0.04~\si{m/s} & \pi/2~\si{rad} \end{matrix}]$ for another $60~\si{s}$.
\end{enumerate}

\section{Discussion} \label{sec:Discussion}

The results in~\cref{fig:TESTS} show excellent preliminary performance. The Modboat configuration is able to travel in a desired direction in either a head-on (\cref{fig:tests:test1}) or sideways (\cref{fig:tests:test2}) orientation, and even mix velocities in the world frame (\cref{fig:tests:test3}). In a significant stress test, the controller is also able to provide excellent 90 degree turning performance, as evidenced by the trajectory in~\cref{fig:tests:test4}, which shows little overshoot and good direction tracking during both legs of the trajectory. Notably, our controller has a difficult time maintaining a steady orientation (all tests in~\cref{fig:TESTS}), but nevertheless maintains reasonable directional swimming.

\section{Conclusion} \label{sec:conclusion}

In this work we have presented a potential-field based control approach to allow a group of three or more docked Modboats to function as a \textit{holonomic} vehicle, improving on prior work~\cite{Knizhnik2022,KnizhnikTRO} that could only create a steerable vehicle. This method works for arbitrary structures of docked modules and has the potential to be scaled to large structures containing many modules.

Preliminary experimental results have shown that this approach is effective in controlling the velocity and somewhat effective at controlling orientation in a few different testing scenarios. 
The major limitation of this strategy is the need for a fixed transition time outside the main swim cycle, which is necessary to avoid collisions but creates undesirable dynamics and slows down the overall controller response time. Future work will consider a transition strategy that minimizes this impact, as well as solution strategies for the swim cycle that minimize the need for transitions. 

Future work will also consider testing with larger numbers of modules and expanding our evaluation to non-controlled environments, like lakes or rivers. This will stress the ability of our controller to reject temporary disturbances. Developing this approach to work with larger configurations will also necessitate developing a more efficient transition solver that displays better scaling. 


\section*{Acknowledgment}

We thank Dr. M. Ani Hsieh for the use of her instrumented water basin in obtaining all of the testing data.

\bibliographystyle{./bibliography/IEEEtran}
\bibliography{./bibliography/IEEEabrv,./bibliography/nonpaper,./bibliography/references,./bibliography/referencesOther}


\clearpage

\end{document}